\documentclass[twoside]{article}

\usepackage{}
\usepackage[utf8]{inputenc}
\usepackage{amsfonts}
\usepackage{graphicx}
\usepackage[left=2.5cm,right=2.5cm,top=2cm,bottom=3cm]{geometry}
\usepackage{subfigure}
\usepackage{epsfig,graphics}
\usepackage{amssymb}
\usepackage{amsmath}
\usepackage{amsthm}
\usepackage{bm}
\usepackage{url}
\usepackage{float}
\usepackage{multirow}
\usepackage{algorithm}
\usepackage{color}

\usepackage[section]{placeins}

\usepackage{booktabs}
\usepackage{xcolor}
\usepackage{makecell}

\usepackage{makecell}
\usepackage{subfig}
\usepackage{tabularx}

\usepackage{authblk}

\newtheorem{theorem}{Theorem}[section]
\newtheorem{lemma}{Lemma}[section]

\newtheorem{definition}{Definition}[section]

\begin{document}

\title{Normalized Iterative Hard Thresholding for Tensor Recovery}

\author[1]{Li Li}
\author[1]{Yuneng Liang}
\author[1]{Kaijie Zheng}
\author[,1,2]{Jian Lu\thanks{Corresponding authors:
jianlu@szu.edu.cn (J. Lu) }}

\renewcommand*{\Affilfont}{\upshape\itshape\small}
\renewcommand\Authands{ and }

\affil[1]{Shenzhen Key Laboratory of Advanced Machine Learning and Applications, \authorcr
\itshape\small{School of Mathematical Sciences, Shenzhen University, Shenzhen 518060, China.}}
\affil[2]{National Center for Applied Mathematics Shenzhen (NCAMS), Shenzhen 518055, China.}

\date{}

\maketitle
\maketitle
\begin{abstract}
Low-rank recovery builds upon ideas from the theory of compressive sensing, which predicts that sparse signals can be accurately reconstructed from incomplete measurements. Iterative thresholding-type algorithms—particularly the normalized iterative hard thresholding (NIHT) method—have been widely used in compressed sensing (CS) and applied to matrix recovery tasks. In this paper, we propose a tensor extension of NIHT, referred to as TNIHT, for the recovery of low-rank tensors under two widely used tensor decomposition models. This extension enables the effective reconstruction of high-order low-rank tensors from a limited number of linear measurements by leveraging the inherent low-dimensional structure of multi-way data. Specifically, we consider both the CANDECOMP/PARAFAC (CP) rank and the Tucker rank to characterize tensor low-rankness within the TNIHT framework. At the same time, we establish a convergence theorem for the proposed TNIHT method under the tensor restricted isometry property (TRIP), providing theoretical support for its recovery guarantees. Finally, we evaluate the performance of TNIHT through numerical experiments on synthetic, image, and video data, and compare it with several state-of-the-art algorithms.

\end{abstract}


\section{Introduction}
Iterative thresholding-type algorithms are primarily developed as improvements on iterative hard thresholding (IHT) \cite{BD09}, including methods such as normalized iterative hard thresholding (NIHT) \cite{BD10}, hard thresholding pursuit (HTP) \cite{F11} and conjugate gradient IHT (CGIHT) \cite{BTW15}. 
That is widely used in compressed sensing (CS) and low-rank matrix recovery (matrix completion) because of its theoretical guarantee.

Specifically, Blumensat et al. \cite{BD10} studied the so-called NIHT algorithm, which is normalized at gradient updates of each step for sparse signal processing. 
Foucart \cite{F11} proposed CGIHT that combine IHT algorithm and the compressive sampling matching pursuit (CoSaMP) algorithm to find sparse solutions of underdetermined linear systems in compressive sensing (CS). 
Blanchard et al. \cite{BTW15} introduced the conjugate gradient IHT (CGIHT) family of algorithms to efficiently solve compressed sensing, which is designed to combine simple hard thresholding with the conjugate gradient method.

In addition, Foucart \cite{F11} provided the HTP and NHTP algorithm (resp. variation of IHT and NIHT) for sparse vector recovery. They establish that under a specific restricted isometry condition, HTP can recover any sparse vector. Later, Zhao et al. \cite{ZDSC17} developed accelerated HTP algorithms to efficiently address the linear least square problem for sparse signal recovery. 
Meanwhile, Foucart et al. \cite{FR13} introduced the Expander IHT algorithm, specifically designed for recovery over lossless expander graphs. Building on this idea, subsequent works have addressed more intricate signal structures and model-based compressed sensing frameworks \cite{BBC14,BKS21}.
Apart from that, there are other works. 
Ollila extended Huber's criterion to sparse recovery \cite{Oll15a, Oll15b} Yang et al. \cite{YFZW11} proposed a Backtracking-based IHT algorithm. 
Han et al. \cite{HNE15} proposed a modified distributed IHT algorithm to improve upon IHT. 
In 2018, Jiang et al. \cite{JDLZLH18} presented a knowledge-aided NIHT algorithm that incorporates a prior information about the probabilities of nonzero entries, and a recursive version (RKA-NIHT) that dynamically updates these probabilities. Matsumoto et al. \cite{MM24} proposed the normalized binary iterative hard thresholding algorithm for 1-bit compressed sensing (CS). Recently, Zhou et al. \cite{Zhou25} based on iterative half thresholding and introduced a new non-negative regularization term to enhance its performance for sparse recovery.

Beyond sparse signal recovery, the theoretical foundations of CS have also inspired the development of low-rank matrix recovery. 
Specifically, Foucart \cite{F11a} extended HTP to recover row-sparse matrices. 
Blanchard et al. \cite{BCHJ14} extended IHT, NIHT, HTP, NHTP, and CoSaMP from the single measurement vector to the multiple MV setting. 
Wei et al. \cite{{WCCL16}} established theoretical recovery guarantees for IHT applied to low-rank matrix recovery. 
Carpentier and Kim \cite{CK16} introduced an estimator for low-rank matrix recovery, which is derived from the IHT method. 
Li et al. \cite{LYYWWL17} proposed a two-dimensional NIHT (2D-NIHT) algorithm to address the high computational complexity of compressive radar imaging. 
Meanwhile, Foucart and Subramanian \cite{FS19} proposed and analyzed a variation of the IHT algorithm tailored for low-rank matrix recovery under subgaussian rank-one measurement models. 
Following \cite{FS19}, Xia et al. \cite{XZ23} in 2023 introduced a new adaptive iterative hard thresholding (AIHT) algorithm with adaptive step size for low-rank matrix recovery, and provided better convergence rate and sampling complexity.

Motivated by the success of recovering low-dimensional structures using Iterative thresholding-type algorithms from limited data, researchers have turned their attention to problem of recovering low-rank tensors.
Specifically, Rauhut et al. \cite{RSS17} extended IHT method to tensor (TIHT) under various tensor decomposition constraints, including the higher order singular value decomposition (HOSVD), the tensor train decomposition  and the general hierarchical Tucker decomposition. 
Goulart et al. \cite{GF15} proposed a step size selection heuristic to accelerate the convergence of the TIHT algorithm for low-rank tensor recovery. 
Grotheer et al. expanded the stochastic IHT algorithm to reconstruct tensors under the Tucker rank constraint \cite{GLMNQ20} , and also expanded the IHT algorithm under the CP rank constraint \cite{GLMNQ21}. Subsequently, they introduced iterative singular tube hard thresholding (ISTHT) algorithms, including basic, accelerated, and stochastic versions to recover low-tubal-rank tensors based on the t-SVD decomposition \cite{GLMNQ24}.  
And they also provided theoretical guarantees using tensor RIP (TRIP). 
The IHT-based algorithms are not only applied to tensor recovery problems, but also to other problems such as phase retrieval \cite{LTW18} and image denoising \cite{SHE16, KHLY18}.

The thresholding-based algorithms for tensor recovery offer several advantages due to their simplicity, computational efficiency, and effectiveness in handling low-rank constraints. These methods iteratively enforce sparsity or low-rank structure by applying thresholding operators, which significantly reduce the computational burden compared to convex relaxation methods. 
Building on these strengths, we propose a novel method in this work 
, called the tensor Normalized Iterative Hard Thresholding (TNIHT) algorithm, aimed at recovering tensors with CP or Tucker rank constraints.

The rest sections of this paper are structured in the subsequent order.
In Section 2, we provide a concise overview of essential concepts and terminologies employed in the tensor structure.
We construct tensor recovery model with incorporating constraints on CP or Tucker rank, and introduce the TNIHT method in Section 3.
In addition, we demonstrate linearly convergent guarantee for the proposed TSVRG method in Section 4.
Section 5 presents all computational outcomes that demonstrate the algorithm's effectiveness by using both synthetic and real data.
Finally, we summarize this work and discuss the future work in Section 6.

\section{Preliminaries and Backgrounds}
    In this section, we provide a concise overview of some fundamental concepts and definitions,
    see \cite{KB09} for more detailed introduction about tensor.
    \subsection{Notations and Operations}
    In this paper, scalars, vectors, matrices and tensors are denoted by lower letter $x\in \mathbb{R}$, bold lower letter x$\in \mathbb{R}^I$, bold capital letter X$\in \mathbb{R}^{n_1\times n_2}$ and calligraphic letter $\mathcal{X}$, respectively.
    A tensor possessed a number of dimensions denoted by $order$.
    We refer to a tensor of the $d$-th order as $\mathcal{X}\in \mathbb{R}^{n_1\times n_2\times \cdots \times n_d}$.
    And each dimension is a $mode$ of $\mathcal{X}$.
    Obviously, we can denote vectors and matrices as tensor with the order $d=1$ and 2, respectively.
    The $i$-th entry of a vector x is represented by x$(i)$. Denote $(i,j)$-th entry of a matrix X. The  $(i_1\times i_2\times \cdots \times i_d)$-th entry of a $d$ order tensor $\mathcal{X}$ is signified by $\mathcal{X}(i_1, i_2, \cdots, i_d)$, where $i_m\in\{1,2,\cdots,n_d\}$ and $m\in\{1,2,\cdots,d\}$.

    Let $\mathcal{X}^{(i)}\in \mathbb{R}^{n_i\times(n_1n_2\cdots n_{i-1}n_{i+1}\cdots n_d)}$ be the representation of $mode$-$i~unfolding$ (unfolding, a.k.a, matricization or flattening)
    which is generated by fixing the $i$-th component of a tensor $\mathcal{X}\in \mathbb{R}^{n_1\times n_2\times \cdots \times n_d}$.

     A $mode$-$i~product$ between a tensor $\mathcal{X}\in \mathbb{R}^{n_1\times n_2\times \cdots \times n_d}$ and a matrix/vector $U\in \mathbb{R}^{n_i\times m_i}$ is represented by $\mathcal{Y}=\mathcal{X}_{\times i}U^T\in \mathbb{R}^{n_1\times n_2\times \cdots \times n_{i-1}\times m_i \times n_{i+1}\times\cdots \times n_d}$ and $\mathcal{Y}^{(i)}=U^T \mathcal{X}^{(i)}$.

    Let $vec(\cdot)$ as an operator employed to vectorize a matrix or a tensor. Analogous to the matrix case, vectorizing a tensor yields a column vector comprising all of elements of the tensor.
    We remain ordering of elements is consistent , i.e $vec(\mathcal{X})=vec(\mathcal{X}^{(1)})$.
    The $inner~product$ of two tensors $\mathcal{X}_1,~\mathcal{X}_2 \in \mathbb{R}^{n_1\times n_2\times \cdots \times n_d}$ is defined to be $$\langle \mathcal{X}_1,\mathcal{X}_2 \rangle \triangleq vec(\mathcal{X}_1)^{T}vec(\mathcal{X}_2).$$
    The induced $Frobenius~norm$ is specified as $$\|\mathcal{X}\|_F\triangleq\sqrt{\langle \mathcal{X},\mathcal{X} \rangle}.$$
  $P_{\Omega}$ represents the corresponding sampling operator that retrieves solely the entries indexed by $\Omega$: 

   \begin{equation*}
   (\mathcal{P}_{\Omega}(\mathcal{X}))=
   \begin{cases}

    \mathcal{X}(i_1, i_2, \cdots, i_d), & \text{if}~  (i_1, i_2, \cdots, i_d)\in \Omega\\
    0   ,                             & \text{if}~  (i_1, i_2, \cdots, i_d)\in \Omega ^c,
   \end{cases}
   \end{equation*}
   where $\Omega ^c$ is the complement of $\Omega$ and $\mathcal{P}_{\Omega}(\mathcal{X})+\mathcal{P}_{\Omega^c}(\mathcal{X})=\mathcal{X}$.
 $\Omega \in \mathbb{R}^{n_1\times n_2\times \cdots \times n_d}$ is a binary index set: $\Omega(i_1, i_2, \cdots, i_d)=1$ if $\mathcal{X}(i_1, i_2, \cdots, i_d)$ is observed, and $\Omega(i_1, i_2, \cdots, i_d)=0$ otherwise.

\subsection{CP and Tucker Decomposition}
$CP~decomposition~and~CP~rank$.
 The cannonical polyadic (CP) decomposition aims to represent the $d$ order tensor
   $\mathcal{X}\in \mathbb{R}^{n_1\times n_2\times \cdots \times n_d}$
   by a the weighted summation of a set of rank-one tensors:
  \begin{equation}\label{CP}
\begin{aligned}
  \mathcal{X}&=\sum\limits_{r=1}^{R}w_r u_r^{(1)}\circ u_r^{(2)}\circ\cdots\circ u_r^{(d)}\\
  &=\mathcal{W}_{\times1}{U_1}\cdots _{\times d}{U_d}
   \end{aligned}
    \end{equation}
    where each rank-one tensor is denoted by the out product \cite{KB09} 
    of $d$ vectors and each $u_r^{(n)},~n=1,2,\cdots,d$ is unit vector with the weight vector w=$[w_1,w_2,\cdots w_r,\cdots w_R]^T\in \mathbb{R}^R$.
    $R$ as the $CP~rank$ of tensor $\mathcal{X}$ is defined as the minimum number of rank-one tensor.

$Tucker~decomposition~and~Tucker~rank$. A tensor $\mathcal{X}\in \mathbb{R}^{n_1\times n_2\times \cdots \times n_d}$ can be symbolized as a core tensor with factor matrices in Tucker decomposition model \cite{KB09}:
 \begin{equation}\label{Tucker}
\begin{aligned}
\mathcal{X}=\mathcal{S}_{\times1}{U_1}\cdots _{\times d}{U_d}.
   \end{aligned}
    \end{equation}
Here, the tensor $\mathcal{S}\in\mathbb{R}^{r_1\times r_2\times \cdots \times r_d}$ is called $core~tensor$ with smaller dimension.
The operator$_{\times i}$ denotes the mode-$i$ product of the tensor.
    The factor matrix with orthonormal columns $U_i\in\mathbb{R}^{n_i\times r_i}$ is the principal components in the $mode$-$i$ for $i=1,2\cdots d$ and $n_i > r_i$.
The $Tucker~rank$ of a $d$ order tensor $\mathcal{X}$ is an $d$ dimension vector, represented as $r=(r_1, \cdots,r_d)$, whose $i$-th element $r_i$ is the rank of mode-$i$ unfolded matrix $\mathcal{X}^{(i)}$ of $\mathcal{X}$. And $r_i$ is also called mode-$i$ rank
with $r_i=rank(\mathcal{X}^{(i)})$.

    Clearly, CP decomposition (\ref{CP}) is a special decomposition case of Tucker decomposition (\ref{Tucker}).
    Meanwhile, CP decomposition (\ref{CP}) can be reformulated as Tucker decomposition (\ref{Tucker}) when the core tensor $\mathcal{W}$ is a super diagonal tensor, i.e., $\mathcal{W}(r,r,\cdots,r)=w_r$.


\subsection{Tensor Recovery Models}
We begin with the the widely recognized low rank tensor recovery problem.
Mathematically, a unified low rank tensor recovery framework can be expressed as
\begin{equation}\label{LRTC}
\begin{array}{cl}
\min\limits_{\mathcal{X}\in \mathbb{R}^{n_1\times n_2\times \cdots \times n_d}} &
{\rm rank}(\mathcal{X})\\
\text{s.t.}&\mathcal{P}_{\Omega}(\mathcal{X})=\mathcal{P}_{\Omega}(\mathcal{T}),
\end{array}
\end{equation}
which reconstructs the missing elements in tensor $\mathcal{T}$ using its partially known ones, the locations of the known entries is given by the observation set $\Omega$.
The ${\rm rank}(\mathcal{X})$ represents the rank of tensor $\mathcal{X}$, and there are various types of tensor ranks.
The missing entries of $\mathcal{X}$ are reconstructed such that the rank of $\mathcal{X}$ is as small as possible based on the assumption that real-world data are structured and inner-correlated.
The optimization problem (\ref{LRTC}) for rank minimization is NP-hard.
The one popular method is tensor nuclear norm minimization as a convex surrogate for rank minimization problem (\ref{LRTC}) as following:
\begin{equation}\label{LRTC_TNN}
\begin{array}{cl}
\min\limits_{\mathcal{X}\in \mathbb{R}^{n_1\times n_2\times \cdots \times n_d}} & \|\mathcal{X}\|_\ast\\
\text{s.t.}&\mathcal{P}_{\Omega}(\mathcal{X})=\mathcal{P}_{\Omega}(\mathcal{T}).
\end{array}
\end{equation}
The data fitting constraint $\mathcal{P}_{\Omega}(\mathcal{X})=\mathcal{P}_{\Omega}(\mathcal{T})$
implies $\mathcal{X}(i_1, i_2, \cdots, i_d)=\mathcal{T}(i_1, i_2, \cdots, i_d)$ for $\{(i_1, i_2, \cdots, i_d)\}\subseteq \Omega$.
Another group of tensor recovery methods involves factorization-based approaches which alternatively update the factor \cite{LAAW20, LZLL21} as following:
\begin{equation}\label{LRTCM_TR}
\begin{array}{cl}
\min\limits_{\mathcal{X}\in \mathbb{R}^{n_1\times n_2\times \cdots \times n_d}} & \frac{1}{2}\|P_{\Omega}(\mathcal{X})-P_{\Omega}(\mathcal{T})\|_F^2\\
\text{s.t.}&{\rm rank}(\mathcal{X})=K.
\end{array}
\end{equation}
$K$ denotes the given bound constraint of low rank tensor $\mathcal{X}$, including CP decomposition, Tucker decomposition and so on.
The differences in tensor recovery brought by various tensor decomposition are mainly reflected in the definition of rank.
Considering solution of the factorization based model (\ref{LRTCM_TR}), one kinds of solution is based on alternating least squares minimization (ALS)\cite{LAAW20,WAA17}.
Unknown data (the least square solution) can be easily obtained by minimizing the sum of squares of the errors between the obtained data and the actual data and approximated step by step in an iterative way.
The other solution is based Riemannian structure \cite{DDLL23,YCZWZ18}.

In addition, a generalized tensor recovery model is quite popular \cite{LLCZ19}, which possesses a predefined rank \cite{GLMNQ20,GLMNQ21,GLMNQ23}.
The optimization problem (\ref{LRTCM_TR}) with the predefined rank $r$ is reformulated as:
\begin{equation}\label{LRTCM_PTR}
\begin{array}{cl}
\min\limits_{\mathcal{X}\in \mathbb{R}^{n_1\times n_2\times \cdots \times n_d}}  & \frac{1}{2}\|y-\mathcal{A}(\mathcal{X})\|_F^2\\
\text{s.t.}&{\rm rank}(\mathcal{X})= r.
\end{array}
\end{equation}
$y\in \mathbb{R}^m$ denotes the observed data.
Assuming that 
$\mathcal{X}^{\ast}$ is the underlying target low rank tensor. The linear map $\mathcal{A}: \mathbb{R}^{n_1\times n_2\times \cdots \times n_d}\rightarrow \mathbb{R}^{m}$ satisfies $y=\mathcal{A}(\mathcal{X}^{\ast})$ to get $\mathcal{X}^{\ast}$.

In the rest of this paper, the rank $r$ of tensor that we are going to talk about CP rank and Tucker rank in recovery model (\ref{LRTCM_PTR}).
In the subsequent section, we present the recovery problem statement and the tensor SVRG algorithm.

\section{Problem Formulation and Tensor NIHT Algorithm}\label{PF}
In this section, we perform the recovery issue of a rank-$r$ tensor $\mathcal{X}^{\ast}$ from $y=\mathcal{A}(\mathcal{X}^{\ast})$.
Specifically, the $i$-th element of $y$ is given as
\begin{equation}
y_i=\mathcal{A}_i(\mathcal{X}^{\ast})=\langle\mathcal{A}_i,\mathcal{X}^{\ast}\rangle,~i\in[m],
\end{equation}
where $[p]=\{1,2,\ldots p\}$ for any positive integer $p$,~$\mathcal{A}_i\in\mathbb{R}^{n_1\times n_2\times \cdots \times n_d}$ is a sensing tensor. Then the cost function $F(x)$ may be constructed as
\begin{equation*}\label{costFunction}
\begin{aligned}
F(\mathcal{X})&\triangleq\frac{1}{m}\|y-\mathcal {A}(\mathcal{X})\|_{2}^{2}
=\frac{1}{m}\sum\limits_{i=1}^{m}(y_i-\langle \mathcal{A}_i,\mathcal{X}\rangle)^{2}\\
&=\frac{1}{M}\sum\limits_{i=1}^{M}(\frac{1}{l}\sum\limits_{j=(i-1)l+1}^{il}((y_j-\langle \mathcal{A}_j,\mathcal{X}\rangle)^{2})\\
&=\frac{1}{M}\sum\limits_{i=1}^{M}\frac{1}{l}\|y_{l_i}-\mathcal{A}_{l_i}(\mathcal{X})\|_{2}^{2}
\triangleq\frac{1}{M}\sum\limits_{i=1}^{M}f_i(\mathcal{X}).
\end{aligned}
\end{equation*}
Where the measurement vector $y\in \mathbb{R}^m$ can be separated into $M$ non-overlapping vector $y_{l_i}\in \mathbb{R}^l,~i\in [M]$. Let $l$ as an integer and $M=\lceil \frac{m}{l} \rceil$. Denote linear operator $\mathcal{A}_{l_i}:\mathbb{R}^{n_1\times n_2\times \cdots \times n_d}\rightarrow\mathbb{R}^l$ with the $j$-th component of $\mathcal{A}_{l_i}(\mathcal{X})$ being $\langle\mathcal{A}_{(i-1)l+j},~\mathcal{X}\rangle,~j\in[l]$.
The function $f_i(\mathcal{X})$ correlates to a collection of measurements $y_{l_i}$.

Next, to recover $\mathcal{X}^{\ast}$, we can think about the following equivalent problem based on the low-rank property of $\mathcal{X}^{\ast}$.
\begin{equation}\label{l0formulation}
\begin{array}{cl}
\min\limits_{\mathcal{X}\in \mathbb{R}^{n_1\times n_2\times \cdots \times n_d}} &F(\mathcal{X})  \\
\text{s.t.}&{\rm rank}(\mathcal{X})= r.
\end{array}
\end{equation}
In order to solve the minimization problem (\ref{l0formulation}),
inspired by the idea \cite{GLMNQ20,GLMNQ21,GLMNQ23} of the authors who promoted the IHT method to the TIHT in addressing recovery issue of low rank tensor with varying rank constraint.
The TIHT is based on GD, and it updates the iterations by
\begin{equation}\label{GD}
\begin{array}{cl}
\mathcal{X}_{t+1}
=\mathcal{X}_{t}-\eta_{t}\nabla F(\mathcal{X}_{t})
=\mathcal{X}_{t}-\frac{\eta_{t}}{M}\sum\limits_{i=1}^{M}\nabla f_i(\mathcal{X}_{t}),
\end{array}
\end{equation}
where $\eta_{t}$ is the stepsize.
Obviously, GD iterations in (\ref{GD}) needs computing of $M$ derivatives, which is computationally expensive.
The natural idea that we randomly select some sample from $\{1,2,\ldots,M\}$ to made up of sample set $l_t$ of size $|l_t|$. This means SGD updates the iterations by
\begin{equation}\label{SGD}
\begin{array}{cl}
\mathcal{X}_{t+1}
=\mathcal{X}_{t}-\frac{\eta_{t}} {|l_t |}\sum\limits_{i \in l_t} \nabla f_i(\mathcal{X}_{t}).
\end{array}
\end{equation}
Obviously, SGD have less the computational cost than GD by comparing (\ref{GD}) to (\ref{SGD}).
Meanwhile, SGD introduces variance due to procedure of random selection.
As a remedy to this issue, we use SVRG \cite{JZ13} that is a variant of SGD, and it can reduce the variance generated by the stochastic process and accelerate convergence rate in this paper,
which updates the iterations in the inner loop by completing a two-step process at the $t$-th iteration:
\begin{equation}\label{IHT}
\begin{array}{cl}
\tilde{\mathcal{X}}_{t}
&=\mathcal{X}_{t}-\eta(\nabla f_{l_{t}}(\mathcal{X}_{t})-\nabla f_{l_{t}}(\widetilde{\mathcal{X}}_{k})+g_{k}) ,\\
\mathcal{X}_{t+1}&=\mathcal{H}_r(\tilde{\mathcal{X}}_{t}).
\end{array}
\end{equation}
Here $g_{k}=\frac{1}{M}\sum\limits_{i=1}^{M}\nabla f_i(\widetilde{\mathcal{X}}_{k})$.
$\tilde{\mathcal{X}}_{k}$ indicates the result of the last iteration update in the outer loop.
Note that the $t$-th iteration (\ref{IHT}) is not direct in the second step.
More detailed process about the solution $\tilde{\mathcal{X}}_{t}$ of the first step, it has to be projected onto the constraint space $\mathcal{M}_{r}$ by employing hard thresholding operator $\mathcal {H}_{r}(\cdot)$, where $\mathcal{M}_{r}=\{\mathcal{X} \in \mathbb{R}^{n_1\times n_2\times \cdots \times n_d}: {\rm rank}(\mathcal{X})=r\}$.
Here the operator $\mathcal{H}_r(\mathcal{X})$ calculates a best rank-$r$ approximation of a tensor $\mathcal{X}$.
We also assume that
\begin{equation}\label{ass}
\begin{array}{cl}
\|\mathcal{H}_r(\tilde{\mathcal{X}}_{t})-\tilde{\mathcal{X}}_{t}\|_F\leq \theta \|\tilde{\mathcal{X}}^{best}_{t} -\tilde{\mathcal{X}}_{t}\|_F
\end{array}
\end{equation}
holds for all $t=1,2,\ldots,n$ with some $\theta\in[1,\infty)$ as same as \cite{RSS17}.
Denote $\tilde{\mathcal{X}}^{best}_{t}={\rm argmin}_{{\rm rank}(\mathcal{X})\leq r}\|\tilde{\mathcal{X}}_{t}-\mathcal{X}\|_F$ as the best rank-$r$ approximation of $\tilde{\mathcal{X}}_{t}$ corresponding with CP or Tucker rank of tensor. 
In our work, we will also presume that such an approximation $\tilde{\mathcal{X}}^{best}_{t}$ is existent.

Based on the previously stated, we introduce tensor stochastic variance reduced gradient method (TSVRG) for tensor recovery problem with CP or Tucker rank constraint as following:
 \begin{center}
     \begin{tabular}{lp{130mm}}
      \hline
      &{Algorithm~$1$~Tensor Stochastic Variance Reduced Gradient~(TSVRG)}\\ 
      \hline
      &{\bf Input}: $r$, $\eta$, $K$, $n$, \\
      &{\bf Output}: $\widehat{\mathcal{X}}=\widetilde{\mathcal{X}}_{k}$\\
      &{\bf Initialize}: $\widetilde{\mathcal{X}}_{0}$\\
      &{\bf for}~~~~$k=0,1,\ldots,K-1$ {\bf do}\\
      &~~~~~~~~~$g_{k}=\frac{1}{M}\sum\limits_{i=1}^{M}\nabla f_i(\widetilde{\mathcal{X}}_{k})$\\
      &~~~~~~~~~$\mathcal{X}_{0}=\widetilde{\mathcal{X}}_{k}$\\
      &~~~~~~~~~{\bf for}~~~~$t=0,\ldots,n-1$~~ {\bf do}\\
      &~~~~~~~~~~~~~~~~~~Randomly select $l_{t}\in\{1,\ldots,M\}$\\ 
      &~~~~~~~~~~~~~~~~~~$\tilde{\mathcal{X}}_{t}=\mathcal{X}_{t}-\eta(\nabla f_{l_{t}}(\mathcal{X}_{t})-\nabla f_{l_{t}}(\widetilde{\mathcal{X}}_{k})+g_{k})$\\
      &~~~~~~~~~~~~~~~~~~$\mathcal{X}_{t+1}=\mathcal {H}_{r}(\tilde{\mathcal{X}}_{t})$\\
      &~~~~~~~~~{\bf end for}\\
      &~~~~~~~~~~$\widetilde{\mathcal{X}}_{k+1}=\mathcal{X}_{n}$ \\
      &~~~~~~~~~~Exit when the stopping criterion is satisfied.\\
      &{\bf end for}\\
     \hline
     \end{tabular}
\end{center}
In above Algorithm, a full gradient $g_{k}$ will be computed in the outer cycle.
The aim is to reduce the variance generated by stochastic process of gradient descent.
After the calculation process passes into the inner cycle.
Firstly, we choose randomly an index set $l_{t}$ from the set $\{1,\ldots,M\}$ and then calculate the SVRG corresponding to the chosen index set.
Here $\mathbb{E}(\nabla f_{l_{t}}(\mathcal{X}_{t}))=g_{k}$ in the inner cycle.
Let gradient as $\nabla f_{l_{t}}(\mathcal{X}_{t})-(\nabla f_{l_{t}}(\widetilde{\mathcal{X}}_{k})-g_{k})$ to gain unbiased gradient and then navigate  the solution along the direction of the gradient to arrive solution $\tilde{\mathcal{X}}_{t}$.
The detailed convergence theory is presented the next section.

Now, we have completed the convergence proof of the Tensor SVRG algorithm.
As we all know, numerous practical applications are closely related to tensors.
In the next section, we focus on two types of tensor data, that are selected for clarity of presentation, and with the aim of confirming our theoretical results.
Specifically, we will examine tensors derived from synthetic data and video data.
The following numerical experiments are designed to demonstrate that TSVRG can indeed be effectively utilized in tensor recovery problem.

\section{Numerical experiments}
The tensor recovery is commonly regarded as a classic rank minimization problem, where the core challenge lies in effectively recovering an approximate form of the original tensor with as low rank as possible from partially observed data.
In this section, we present some numerical experiments to demonstrate the performance of TSVRG method for tensor recovery task with CP or Tucker rank constraint.
All experiments adopt third-order tensor data (i.e., $d=3$), including synthetic and real data.
For comparisons of comprehensiveness and completion, we first contrast recovery performance based on CP rank restraint of tensor with TIHT \cite{GLMNQ21}.
Then we contrast recovery performance based on Tucker rank restraint of tensor with TIHT and StoIHT \cite{GLMNQ20}.
In these methods, TIHT is based on GD, and StoIHT or TSVRG is based on SGD. And TSVRG is designed to reduce the variance generated in SGD process, while helps to converge faster than standard SGD.
The red, green and blue lines in figures of numerical results represent TIHT, StoTIHT and TSVRG method, respectively.
In all the existing methods, numerical results will verify that TSVRG performs the best with CP rank constraint or low Tucker constraint both synthetic and real data.
In addition, we employ Barzilai-Borwein (BB) \cite{BB88,TMDQ16} technique for automatic determination of step sizes $\eta$.
And the $k$-th iteration, we adopt $\eta_{k}=\|\widetilde{\mathcal{X}}_{k}-\widetilde{\mathcal{X}}_{k-1}\|_{F}^{2}
/\langle\widetilde{\mathcal{X}}_{k}-\widetilde{\mathcal{X}}_{k-1},g_{k}-g_{k-1}\rangle$.
All the methods are executed on desktop of Window 11 and Matlab (R2023a) with Intel(R) Core(TM) i7-10700 CPU at 2.90 GHz  and 64 GB RAM.

In terms of parameter setting, we establish the maximum number of iterations K =500 both synthetic data and real data for all methods.
Algorithm 1 will stop if $\text{RSE}\leq10^{-3}$, which signifies that the algorithm has converged; or it will stop if the number of iterations reached K.
We randomly select $30\%, 50\%, 70\%, 90\%$ entries of each tensor as missing data,
and employ the abbreviation “SR" to signify Sample Ratio.

At the same time, we select the data (synthetic and video data) and other parameters (such as CP or Tucker rank $r$,  corresponding sample ratio SR and so on) of the compared algorithms according to guidance provided in the source papers.
We assess the performance of tensor recovery using two common criteria.
Namely the Relative Square Error (RSE) and the Peak Signal-to-Noise Ratio (PSNR), it can be defined by
\begin{equation*}
\begin{aligned}
\text{RSE}:=\frac{\|\mathcal{X}^{\ast}-\mathcal{X}\|_F}{\|\mathcal{X}\|_F},
~~~~~\text{PSNR}:=10log_{10}(\frac{\|\mathcal{X}\|^2_\infty}{\frac{1}{i_1 i_2 \cdots  i_d}\|\mathcal{X}^{\ast}-\mathcal{X}\|^2_F}).
\end{aligned}
\end{equation*}
Where $\mathcal{X}^{\ast}$ represents the recovered tensor based on a few entries from the (ground truth) $\mathcal{X}$.
In each figure of numerical experiment, we also observe varying levels of of what is referred to as the “convergence horizon”.
In other words, convergence horizon is the RSE values after the algorithm has reached a relative and stable state in operation process.
It reflects that the precision level (corresponds to the effect of tensor recovery problem) of the algorithm can achieve after a certain number of iterations.
A smaller RSE value indicates a better recovery effect.
Another metric for assessing the recovery effect is PSNR. It is a commonly used image quality assessment indicator that measures the similarity between the recovered image and the original image. A higher PSNR value indicates better visual recovery quality (higher resolution) of the image.
We repeat each run 5 times and report the average results.
We also show the execution time of the each algorithm in real data, in seconds.
For the sake of simplicity in presentation, we have rounded the running time and the PSNR to two decimal places.

\subsection{Performance comparison with CP rank restraint}
In this subsection, we demonstrate the effectiveness of tensor recovery comparing with TIHT in synthetic and real data under CP rank restraint. 

\subsubsection{Virtual}
Firstly, for verification of real data, we produce $n_{1}\times n_{2}\times n_{3}$ tensors of CP rank $r$, whose elements adhere to the the Gaussian distributions.
The positions of observed entries over $n_{1}\times n_{2}\times n_{3}$ are randomly and uniformly selected.
And we set the size of tensor to be $n_{1}=n_{2}=n_{3}=10$, the different sample ratio SR = $\{30\%, 50\%, 70\%, 90\%\}$ and CP rank $r$ from 2 to 5 for demonstrating the performance comparing with TIHT for tensor recovery.
Figure 1 presents outcomes that illustrate the RSE of different ranks as well as different sample ratio.
In our first experiment, we report the results of RSE of TIHT and TSVRG method
versus varying CP rank $r$ with SR=0.5 in the left of Figure 1.
To further validate the effectiveness of TSVRG, we adopt CP rank $r=2$ and different sampling ratio, and show the recovery ability by RSE of TIHT and TSVRG method versus different sampling ratio in the right of Figure 1.
The difference in initial values arises from the random picking of the observed location.
And we can find the similar phenomena in the later experiments.

As shown on the left in Figure 1, as expected, both TIHT and TSVRG method demonstrate that the lower CP rank display more rapid convergence tendency.
However, this does not imply a slower tendency of convergence for tensors with larger CP rank. For example, convergence rate of $r=5$ is more effective than one of $r=4$ in terms of TSVRG method.
The same performance appears in TIHT (convergence rate of $r=5$ is better than one of $r=3$ and 4).
In addition, the convergence rate of the two methods indicates that TSVRG outperforms TIHT under the same rank conditions.
The last and most important, the convergence horizon and the convergence rate of TSVRG are superior to those of TIHT in all the demonstrated ranks.

And then, we observe in the Figure 1 on the right, again as predicted, that lager sample ratio enjoy quicker convergence tendency.
As shown, the convergence horizon of TSVRG exhibits a higher level of performance compared to  one of TIHT in all values of the varing sampling ratio.
And the convergence rate of TSVRG demonstrates outperforms than one of TIHT under each different sample ratio.
Furthermore, about the convergence rate, TSVRG in every sampling ratio outshines TIHT in SR=$30\%, 50\%, 70\%$, but TIHT in SR=$90\%$ is higher than TSVRG in SR=$30\%, 50\%$.
On the whole, the recovery performance of TSVRG is better than that of TIHT in synthetic data under CP rank restraint.

\subsubsection{Image}

\subsubsection{Video}

Next, in order to verify the recovery effectiveness of the TSVRG algorithm in real data under CP rank restraint.
We select the real tensor data (Candle video) using the original paper.
For sake of computation, we use smaller image that the size of the Candle video is $30\times30\times10$. Where the 10 images are a part of of larger image.
The position of observed entries are produced randomly, and the CP rank and SR is 15 and 0.8, respectively.
Here, Both the PSNR and RSE are employed for the comparison.
As shown on top left in Figure 2, the quantitative comparisons show that TSVRG can provide faster convergence rate and lower convergence horizon than ones of TIHT.
In addition, We discover that the proposed TSVRG algorithm escapes from local minima 3 times within 100 iterations, which also validates the theoretical advantage of SVRG over GD or SGD in numerical experiments.
We also show the 150th iteration that the execution time of TIHT and TSVRG is 304.05s and 161.30s; PSVR of TIHT and TSVRG is 25.70dB and 28.76dB. 
That is to say, as shown, the recovered first frame in 150 iterations by TSVRG has of visualization higher quality in comparison with TIHT.
The experimental results verify that our method is still comparable to TIHT with respect to synthetic and real data under CP rank restraint.

Now that we have completed the effectiveness validation of our proposed TSVRG method under CP rank restraint, we will proceed to continue with the performance verification of TSVRG under Tucker rank restraint for tensor recovery.

\subsection{Performance comparison with Tucker rank restraint}
In this subsection, we show the efficacy of tensor recovery when compared to TIHT and StoTIHT in synthetic and real data under Tucker rank restraint.

\subsubsection{Virtual}

Likely, first for verification of real data,
we produce $n_{1}\times n_{2}\times n_{3}$ tensors of Tucker rank $r$.
And its entries always keep to the the Gaussian distributions.
The positions of observed entries over $n_{1}\times n_{2}\times n_{3}$ are also randomly and uniformly selected.
We investigate the size of tensor to be $n_{1}=n_{2}=5,~n_{3}=6$, the four sample ratio SR=$\{30\%, 50\%, 70\%, 90\%\}$ and four different Tucker rank $r=(1,1,2;1,2,2;2,2,2;2,2,3)$
to validate the effectiveness of tensor recovery by comparing TSVRG with TIHT and StoTIHT.
As shown in Figure 3, we illustrate the RSE versus different Tucker ranks with SR = 0.5 on the left, and different sample ratio with Tucker rank $r=(1,2,2)$ on the right for three tensor recovery methods.

In the first experiment on the left of the Figure 3, on the whole, the outcomes show that convergence rate of TSVRG are superior to those of TIHT and StoTIHT.
Except for convergence rate of TIHT with Tucker rank $r=(1,1,2)$ is better than one of TSVRG with $r=(1,2,2)$ and $r=(2,2,3)$.
In addition, the convergence rate and the convergence horizon of the three methods indicate that TSVRG outperforms TIHT and StoTIHT under the same Tucker rank conditions.
We observe similar trend in the Figure on the right that larger sample ratio will generate quicker convergence rate.
At the same time, the convergence rate TSVRG exhibits a higher level performance compared to  one of TIHT and StoTIHT under each same sampling ratio.
Overall, the convergence rate of our TSVRG method is evidently superior to those of TIHT and StoTIHT; and the convergence horizonin of TSVRG is slightly better than other two method.
So, TSVRG holds obvious advantageous in comparison with other two algorithms in synthetic data under Tucker rank restraint.


\subsubsection{Image}



\subsubsection{Video}

Moving forward, to confirm the recovery efficacy of the TSVRG algorithm in real data under Tucker rank restraint.
Finally, we adopt recovery effect of video data to compare three varing methods.
Same as case of CP rank restraint, we select the real tensor data (Candle video) using the original paper. And we choice size of the Candle video and way of observed position is also the same as the case of CP rank restraint.
In addition, we adopt the Tucker rank and SR is $r=(8.8.2)$ and SR=0.5, respectively.
For this comparison with TIHT and StoTIHT, we also have applied both PSNR and RSE as our metrics.
As shown the top left of Figure 4, the quantitative comparisons demonstrate that TSVRG generating lower convergence horizon within limited iterations and faster convergence rate than ones of TIHT and StoTIHT.

To further demonstrate the effectiveness of TSVRG method, we present the recovered outcomes of first frame in the 20 iteration by different methods in Figure 4.
And we present in 20th iteration that the execution time of TIHT StoTIHT TSVRG algorithms are 0.08s, 0.03s and 0.06s, respectively; and respective PSNR of the three algorithms are 22.56dB, 21.20dB, and 23.04dB.
It can be seen that that our algorithm achieves the highest PSNR with slightly less time among all methods within 20 iterations.
The experimental results verify that TSVRG outperforms TIHT and StoTIHT for both synthetic and real data under Tucker rank restraint.

Without a doubt, these experiments demonstrate that TSVRG algorithm is consistently superior to other tensor recovery algorithms based on GD and SGD under CP or Tucker rank restraint.
At the same time, numerical results reveal the theoretical advantages of the proposed algorithm, specifically the inference that SVRG can escape local minima and accelerate the convergence rate.

\section*{Acknowledgments}
The work is partially supported by the NSF of China under grants U21A20455, 12326619, 61972265, by the NSF of Guangdong Province of China under grants 2020B1515310008 and 2023A1515011691, and by the Educational Commission of Guangdong Province of China under grant 2019KZDZX1007.
At the same time, the authors are also sincerely thank the reviewers for their constructive suggestions that has enhanced the quality of this manuscript.


\begin{thebibliography}{1}

\bibitem{AD19} S. Athisayamani and  D. Dejey, (2019). A novel video coding framework with tensor representation for efficient video streaming. Wireless Personal Communications, 109 (4), pp. 2699-2717.


\bibitem{BBC14} Bubacarr Bah, Luca Baldassarre, Volkan Cevher, Model-based sketching and recovery with expanders, in: Proceedings of the Twenty-Fifth Annual ACM-SIAM Symposium on Discrete Algorithms, SIAM, 2014, pp. 1529–1543.
 
\bibitem{BKS21} Bubacarr Bah, Jannis Kurtz, Oliver Schaudt, Discrete optimization methods for group model selection in compressed sensing, Math. Program. 190 (1) (2021) 171–220.


\bibitem{BB88} J. Barzilai and J. Borwein, (1988). Two-point step size gradient methods. IMA Journal of Numerical Analysis, 8 (1), pp. 141-148.


\bibitem{BCHJ14} J. D. Blanchard, M. Cermak, D. Hanle and Y. Jing, (2014). Greedy algorithms for joint sparse recovery. IEEE Transactions on Signal Processing, 62 (7), pp. 1694-1704.


\bibitem{BTW15} J. D. Blanchard, J. Tanner and K. Wei, (2015). CGIHT: conjugate gradient iterative hard thresholding for compressed sensing and matrix completion. Information and Inference: A Journal of the IMA, 4 (4), pp. 289-327.





\bibitem{BD08} T. Blumensath and M. Davies, (2008). Iterative thresholding for sparse approximations. Journal of Fourier Analysis and Application, 14 (5), pp. 629-654.


\bibitem{BD10} Blumensath, T. , and M. E. Davies . "Normalized Iterative Hard Thresholding: Guaranteed Stability and Performance." IEEE Journal of Selected Topics in Signal Processing 4.2(2010):298-309.


\bibitem{BD09} Blumensath, Thomas, and Mike E. Davies. Iterative hard thresholding for compressed sensing.  Applied and computational harmonic analysis 27.3 (2009): 265-274.




\bibitem{CSMZT20} C. Caiafa, J, Sol$\acute{e}$-Casals, P. Marti-Puig, S. Zhe and T. Tanaka, (2020). Decomposition methods for machine learning with small, incomplete or noisy datasets.  Applied Sciences, 10 (23), DOI: 10.3390/app10238481.




\bibitem{CYLY23} C. Cao, H. Yue, X. Liu and J. Yang, (2023). Unsupervised HDR image and video tone mapping via contrastive learning. arXiv: 2303.07327v2.



\bibitem{CK16} A. Carpentier and A. Kim, (2016). An iterative hard thresholding estimator for low rank matrix recovery with explicit limiting distribution. arXiv preprint arXiv:1502.04654v3 [math.ST].


\bibitem{CYZFZZ20} Y. Chang, L. Yan, X. Zhao, H. Fang, Z. Zhang and S. Zhong, (2020). Weighted low-rank tensor recovery for hyperspectral image restoration. IEEE Transactions on Cybernetics, 50 (11), pp. 4558–4572.


\bibitem{CY25} J. Chen and M. Yuan, (2025). Optimal Quantized Compressed Sensing via Projected Gradient Descent. arXiv preprint arXiv:2407.04951v2.





\bibitem{DDLL23} S. Duan, X. Duan, C. Li and J. Li, (2023). Riemannian conjugate gradient method for low-rank tensor completion. Advances in Computational Mathematics, 49 (3), doi: 10.1007/s10444-023-10036-0.



\bibitem{DES22} D. Driggs, M. Ehrhardt and C. Sch$\ddot{o}$nlieb, (2022). Accelerating variance-reduced stochastic gradient methods. Mathematical Programming, 191, pp. 671–715.


\bibitem{FL18} S. Friedland and L. Lim, (2018). Nuclear norm of higher-order tensors. Mathematics of Computation, 87 (311), pp. 1255–1281.


\bibitem{FRW11} M. Fornasier, H. Rauhut and R. Ward, (2011). Low-rank matrix recovery via iteratively reweighted least squares minimization. SIAM Journal on Optimization, 21, pp. 1614–1640.


\bibitem{F11} S. Foucart, (2011). Hard Thresholding Pursuit: An algorithms for compressive sensing. SIAM J. NUMER. ANAL. Vol. 49, No. 6, pp. 2543–256.



\bibitem{F11a} S. Foucart, (2011). Recovering jointly sparse vectors via hard thresholding pursuit. In 2011 18th IEEE International Conference on Image Processing (ICIP), pp.

\bibitem{FR13} S. Foucart, H. Rauhut, A mathematical introduction to compressive sensing, first ed., Applied and Numerical Harmonic Analysis, 10.1007/978-0-8176-4948-7, Birkhauser New York, NY, 2013, p. XVIII, 625, Copyright: Springer Science and Business Media New York.


\bibitem{FS19} S. Foucart and S. Subramanian, (2019). Iterative hard thresholding for low-rank recovery from rank-one projections. Linear Algebra and its Applications, 572, pp. 117-134.



\bibitem{GF15} J. Goulart and G. Favier, (2015). An iterative hard thresholding algorithm with improved convergence for low-rank tensor recovery. In 2015 23rd European Signal Processing Conference (EUSIPCO), pp. 1666-1670.

\bibitem{GLMNQ20} R. Grotheer, S. Li, A. Ma, D. Needell and J. Qin, (2020). Stochastic iterative hard thresholding for low-tucker-rank tensor recovery. 2020 Information Theory and Applications Workshop (ITA).

\bibitem{GLMNQ21} R. Grotheer, S. Li, A. Ma, D. Needell and J. Qin, (2021). Iterative hard thresholding for low CP-rank tensor models. Linear and Multilinear Algebra, 70 (22), pp. 7452–7468.

\bibitem{GLMNQ23} R. Grotheer, S. Li, A. Ma, D. Needell and J. Qin, (2023). Iterative singular tube hard thresholding algorithms for tensor completion. arXiv: 2304.04860v1.

\bibitem{GLMNQ24} R. Grotheer, S. Li, A. Ma, D. Needell and J. Qin, (2024). Iterative singular tube hard thresholding algorithms for tensor recovery. Inverse Problems and Imaging, 18 (4), pp. 889-907.


\bibitem{GSGHY23} J. Guo, Y. Sun, J. Gao, Y. Hu and B. Yin, (2023). Logarithmic schatten-p norm minimization for tensorial multi-View subspace clustering. IEEE Transactions on Pattern Analysisb and Machine Intelligence, 45(3), pp. 3396–3410.

\bibitem{HNE15} P. Han, R. Niu and Y. C. Eldar, (2015). MODIFIED DISTRIBUTED ITERATIVE HARD THRESHOLDING. In 2015 IEEE International Conference on Acoustics, Speech and Signal Processing (ICASSP), pp. 3766-3770.

\bibitem{HNLN22} N. Han, J. Nie, J. Lu and M. Ng, (2022). Stochastic variance reduced gradient for affine rank minimization problem. arXiv: 2211.02802v1.

\bibitem{HL13} C. Hillar and L. Lim, (2013). Most tensor problems are NP-hard. Journal of the ACM, 60 (6), pp. 1-39.


\bibitem{JDLZLH18} Q. Jiang, R. C. de Lamare, Y. Zakharov, S. Li and X. He, (2018). Knowledge-aided normalized iterative hard thresholding algorithms and applications to sparse reconstruction. arXiv:1809.09281.


\bibitem{JZZM23} T. Jiang, X. Zhao, H. Zhang and K. Michael, (2023). Dictionary learning with low-rank coding coefficients for tensor completion. IEEE Transactions on Neural Networks and Learning Systems, 34 (2), pp. 932-946.


\bibitem{JPTD17} A. Johann, H. Phien, H, Tuan and M. Do, (2017). Efficient tensor completion for color image and video recovery: low-rank tensor train. IEEE Transactions on Image Processing, 26 (5), pp. 2466-2479.


\bibitem{JZ13} R. Johnson and T. Zhang, (2013). Accelerating stochastic gradient descent using predictive variance reduction. NIPS'13: Proceedings of the 26th International Conference on Neural Information Processing Systems, 1, pp. 315–323.

\bibitem{KB09} T. Kolda and B. Bader, (2009). Tensor decompositions and applications. SIAM review, 51, pp. 455-500.

\bibitem{KHLY18} Z. Kong, L. Han, X. Liu and X. Yang, A New 4-D Nonlocal Transform-Domain Filter for 3-D Magnetic Resonance Images Denoising, IEEE Transactions on Medical Imaging, vol. 37, no. 4, pp. 941-954.

\bibitem{LAAW20} X. Liu,  S. Aeron, V. Aggarwal and X. Wang, (2020). Low-tubal-rank tensor completion using alternating minimization. IEEE Transactions on Information Theory, 66 (3), pp. 1714-1737.


\bibitem{LYYWWL17} G. Li, J. Yang, W. Yang, Y. Wang, W. Wang, and L. Liu, (2017). 2D Normalized Iterative Hard Thresholding Algorithm for Fast Compressive Radar Imaging. Remote Sensing, 9(6), 619.

\bibitem{LTW18} S. Li, G. Tang, and M. B. Wakin, (2018). Simultaneous blind deconvolution and phase retrieval with tensor iterative hard thresholding. In 2018 IEEE International Conference on Acoustics, Speech and Signal Processing (ICASSP), pp. 4184-4188.

\bibitem{LLHZ20} Y. Liu, Z. Long, H. Huang and C. Zhu, (2020). Low CP rank and Tucker rank tensor completion for estimating missing components in image data. IEEE Transactions on Circuits and Systems for Video Technology, 30 (4), pp. 944-954.


\bibitem{LMWY13} J. Liu, P. Musialski, P. Wonka and J. Ye, (2013). Tensor completion for estimating missing values in visual data. IEEE Transactions on Pattern Analysis and Machine Intelligence, 35 (1), pp. 208–220.

\bibitem{LXHLJML22} J. Lu, C. Xu, Z. Hu, X. Liu, Q. Jiang, D. Meng and Z. Lin, (2022). A new nonlocal low-rank regularization method with applications to magnetic resonance image denoising. Inverse Problems, 38 065012.


\bibitem{LZL17} Z. Lu, Y. Zhang and J. Lu, (2017). $\ell_p$ Regularized low-rank approximation via iterative
reweighted singular value minimization. Computational Optimization and Applications, 68, pp. 619–42

\bibitem{LLCZ19} Z. Long, Y. Liu, L. Chen and C. Zhu, (2019). Low rank tensor completion for multiway visual data. Signal Processing, 155, pp. 301-316.


\bibitem{LZLL21} Z. Long, C. Zhu, J. Liu and Y. Liu, (2021). Bayesian low rank tensor ring for image recovery. IEEE Transactions on Image Processing, 30, pp. 3568-3580.


\bibitem{MM24} N. Matsumoto and A. Mazumdar. Binary iterative hard thresholding converges with optimal number of measurements for 1-bit compressed sensing. Journal of the ACM, 71(5):1–64, 2024.





\bibitem{MIDDC21} V. Monardo, A. Iyer, S. Donegan, M. De Graef and Y. Chi, (2021). Plug-And-Play image reconstruction meets stochastic variance-reduced gradient methods. 2021 IEEE International Conference on Image Processing (ICIP), pp. 2868-2872.

\bibitem{NNW17} N. Nguyen, D. Needell and T. Woolf, (2017). Linear convergence of stochastic iterative greedy algorithms with sparse constraints. IEEE Transactions on Information Theory, 63 (11), pp. 6869-6895.


\bibitem{Oll15a} E. Ollila, (2015). Multichannel sparse recovery of complex-valued signals using Huber's criterion. arXiv:1504.04184.


\bibitem{Oll15b} E. Ollila, (2015). Nonparametric simultaneous sparse recovery: an application to source localization. arXiv: 1276226.


\bibitem{O11} I. Oseledets, (2011). Tensor-train decomposition. SIAM Journal on Scientific Computing, 33 (5), pp. 2295–2317.

\bibitem{QLNMGHD21} J. Qin, S. Li, D. Needell, A. Ma, R, Grotheer, C. Huang and N. Durgin, (2021). Stochastic greedy algorithms for multiple measurement vectors. Inverse Problems and Imaging, 15 (1), pp. 79-107.





\bibitem{RSS17} H. Rauhut, R. Schneider and $\check{Z}$. Stojanac, (2017). Low-rank tensor recovery via iterative hard thresholding. Linear Algebra and its Applications, 523, pp. 220-262.

\bibitem{BCDH10} Richard G. Baraniuk, Volkan Cevher, Marco F. Duarte, Chinmay Hegde, Model-based compressive sensing, IEEE Trans. Inform. Theory 56 (4) (2010) 1982–2001.


\bibitem{SKM19} H. Sato, H. Kasai and B. Mishra, (2019). Riemannian stochastic variance reduced gradient with retraction and vector transport. SIAM Journal on Optimization, 29 (201), pp. 1444–1472.


\bibitem{SWLLZG22} F. Shang, B. Wei, H. Liu, Y. Liu, P. Zhou and M. Gong, (2022). Efficient gradient support pursuit with less hard thresholding for cardinality-constrained learning. IEEE Transactions on Neural Networks and Learning Systems, 33 (12), pp. 7806-7817.



\bibitem{SCLYL22} Q. Shen, Y. Chen, Y. Liang, S. Yi and W. Liu, (2022). Weighted schatten$p$-norm minimization with logarithmic constraint for subspace clustering. Signal Processing, 198(9):108568. DOI:10.1016/j.sigpro.2022.108568.

\bibitem{SHE16} Yi. Shen, Bin. Han, E. Braverman (2016). Adaptive frame-based color image denoising. Applied and computational Harmonic Analysis, 41(1), 54-74.


\bibitem{SLC17} Q. Shi, H. Lu, and Y. Cheung, Tensor rank estimation and completion via cp-based nuclear norm. \emph{CIKM '17}, (2017), 949-958.


\bibitem{TY10} K. Toh and S. Yun, (2010). An accelerated proximal gradient algorithm for nuclear norm regularized least squares problems. Pacific Journal Of Optimization, 6, pp. 615–640.

\bibitem{TMDQ16} C. Tan, S. Ma, Y. Dai and Y. Qian, (2016). Barzilai-borwein step size for stochastic gradient descent. Advances in Neural Information Processing Systems, (29).

\bibitem{TLZPHJL23} Z. Tu, J. Lu, H. Zhu, H. Pan, W. Hu, Q. Jiang and Z. Lu, (2023). A new nonconvex low-rank tensor approximation method with applications to hyperspectral images denoising. Inverse Problems, 39 065003.


\bibitem{WCH23} Q. Wang, C. Cui and D. Han, (2023). Accelerated doubly stochastic gradient descent for tensor CP decomposition. Journal of Optimization Theory and Applications, https://doi.org/10.1007/s10957-023-02193-5.

\bibitem{WZG17} L. Wang, X. Zhang and Q. Gu, (2017). A unified variance reduction-based framework for nonconvex low-rank matrix recovery. 34th International Conference on Machine Learning, 70.

\bibitem{WAA17} W. Wang, V. Aggarwal and S. Aeron, (2017). Efficient low rank tensor ring completion. 2017 IEEE International Conference on Computer Vision (ICCV), doi: 10.1109/ICCV.2017.607.


\bibitem{WCCL16} K. Wei, J. F. Cai, T. F. Chan and S. Leung, (2016). Guarantees of Riemannian Optimization for Low Rank Matrix Recovery. arXiv preprint arXiv:1511.01562.


\bibitem{XZ23} Y. Xia and L. Zhou, (2023). Adaptive iterative hard thresholding for low-rank matrix recovery and rank-one measurements. Journal of Complexity, 76, 101725.

\bibitem{YLDS21} T. Yu, X. Liu, Y. Dai and J. Sun, (2021). Stochastic variance reduced gradient methods using a trust-region-like scheme. Journal of Scientific Computing, 87.

\bibitem{YLDS22} T. Yu, X. Liu, Y. Dai and J. Sun, (2022). Variable metric proximal stochastic variance reduced gradient methods for nonconvex nonsmooth optimization. Journal of Industrial and Management Optimization, 18 (4), pp. 2611–2631.

\bibitem{YFZW11} H. Yang, H. Fang, C. Zhang and S. Wei, (2011). Iterative hard thresholding algorithm based on backtracking. Acta Automatica Sinica, 37(3), pp. 276-282.

\bibitem{YLWXK19} L. Yang, M. Liu, J. Wang, X. Xie and J. Kuang, (2019). Tensor completion for recovering multichannel audio signal with missing data. China Communications, 16 (4), pp. 186-195.

\bibitem{YCZWZ18} L. Yuan, J. Cao,  X. Zhao,  Q. Wu and Q. Zhao, (2018). Higher-dimension tensor completion via low-rank tensor ring decomposition. 2018 Asia-Pacific Signal and Information Processing Association Annual Summit and Conference (APSIPA ASC), doi: 10.23919/APSIPA.2018.8659708.


\bibitem{YYLHFZ15} L. Yuan, Z. Yu, W. Luo, Y. Hu, L. Feng and A. Zhu, (2015). A hierarchical tensor-based approach to compressing, updating and querying geospatial data, IEEE Transactions on Knowledge and Data Engineering, 27 (2), pp. 312–325.




\bibitem{ZZXZC16} Q. Zhao, G. Zhou, S. Xie, L. Zhang and A. Cichocki, (2016). Tensor ring decomposition. arxiv: 1606.05535.




\bibitem{ZDSC17} X. Zhao, P. Du, T. Sun, and L. Cheng, (2017). Accelerated Hard Thresholding Algorithms for Sparse Recovery. Advances in Engineering Research (AER), 130, pp. 1322-1328.



\bibitem{ZYFWCXY21} S. Zhang, L. Yang, J. Feng, W. Wei, Z. Cui, X. Xie and P. Yan, (2021). A tensor-network-based big data fusion framework for Cyber-Physical-Social Systems (CPSS). Information Fusion, 76 (C), pp. 337–354.

\bibitem{ZZZAXC16} G. Zhou, Q. Zhao, Y. Zhang, T. Adali, S. Xie and A. Cichocki, (2016). Linked component analysis from matrices to high-order tensors: applications to biomedical data. Proceedings of the IEEE, 104 (2), pp. 310–331.

\bibitem{ZLFLY21} P. Zhou, C. Lu, J. Feng, Z. Lin and S. Yan, (2021). Tensor low-rank representation for data recovery and clustering. IEEE Transactions on Pattern Analysis and Machine Intelligence, 43 (5), pp. 1718-1732.

\bibitem{Z18} X. Zhou, (2018). On the fenchel duality between strong convexity and lipschitz continuous gradient. arXiv: 1803.06573.

\bibitem{Zhou25} X. Zhou, Z. Liu, H. Zhang, Z. Zhao, and Y. Liu, (2025). Key parameters for iterative thresholding-type algorithm with nonconvex regularization. Digital Signal Processing, 164, 105246.


\bibitem{ZHZZJ21} Y. Zheng, T. Huang, X. Zhao, Q. Zhao and T. Jiang, (2021). Fully-connected tensor network decomposition and its application to higher-order tensor completion. AAAI Conference on Artificial Intelligence18.

\bibitem{ZHZZJJM20} Y. Zheng, T. Huang, X. Zhao, Q. Zhao, T. Jiang, T. Ji and T. Ma, (2020). Tensor n-tubal rank and its convex relaxation for low-rank tensor completion. Information Science, 532, pp. 170–189.



\end{thebibliography}
\end{document}